\documentclass[10pt,twocolumn,letterpaper]{article}

\usepackage{iccv}              

\definecolor{iccvblue}{rgb}{0.21,0.49,0.74}
\usepackage[pagebackref,breaklinks,colorlinks,allcolors=iccvblue]{hyperref}
\usepackage{array}
 \usepackage{multirow}

\makeatletter
\renewcommand{\@fnsymbol}[1]{\ifcase#1\or \dag\or *\fi}  
\makeatother

 \makeatletter
\newcommand\blfootnote[1]{%
  \begingroup
    \renewcommand\thefootnote{}%
    \footnote{#1}%
    \addtocounter{footnote}{-1}%
  \endgroup
}
\makeatother

\def\paperID{*****} 
\def\confName{ICCV}
\def\confYear{2025}

\title{FusionGen: Feature Fusion-Based Few-Shot EEG Data Generation}

\author{
\textbf{Yuheng Chen}\textsuperscript{1}\thanks{~These authors contributed equally to this work.} \kern0.5em  
\textbf{Dingkun Liu}\textsuperscript{1,2\dag}
\textbf{Xinyao Yang}\textsuperscript{1}\kern0.5em 
\textbf{Xinping Xu}\textsuperscript{1}\kern0.5em 
\textbf{Baicheng Chen}\textsuperscript{1}\kern0.5em 
\textbf{Dongrui Wu}\textsuperscript{1,2}\thanks{~Corresponding author: Dongrui Wu, \texttt{drwu09@gmail.com}}  
\\[0.7em]
\textsuperscript{1}Huazhong University of Science and Technology, Wuhan, 430074, China\\
\textsuperscript{2}Zhongguancun Academy, Beijing, 100094, China\\[0.3em]
\ttfamily chenyuheng@hust.edu.cn \quad liudingkun@hust.edu.cn \quad drwu09@gmail.com
}

\begin{document}
\maketitle

\blfootnote{This research was supported by Open Fund of Intelligent Control Laboratory 2024-ZKSYS-KF02-04 and Zhongguancun Academy 20240301.}
\blfootnote{This paper was published as a conference paper at ICCV 2025.}

\begin{abstract}

Brain-computer interfaces (BCIs) provide potential for applications ranging from medical rehabilitation to cognitive state assessment by establishing direct communication pathways between the brain and external devices via electroencephalography (EEG). However, EEG-based BCIs are severely constrained by data scarcity and significant inter-subject variability, which hinder the generalization and applicability of EEG decoding models in practical settings. To address these challenges, we propose FusionGen, a novel EEG data generation framework based on disentangled representation learning and feature fusion. By integrating features across trials through a feature matching fusion module and combining them with a lightweight feature extraction and reconstruction pipeline, FusionGen ensures both data diversity and trainability under limited data constraints. Extensive experiments on multiple publicly available EEG datasets demonstrate that FusionGen significantly outperforms existing augmentation techniques, yielding notable improvements in classification accuracy. 

\end{abstract}    
\section{Introduction}
\label{sec:intro}

\begin{figure}[t]
  \centering
   \includegraphics[width=0.95\linewidth]{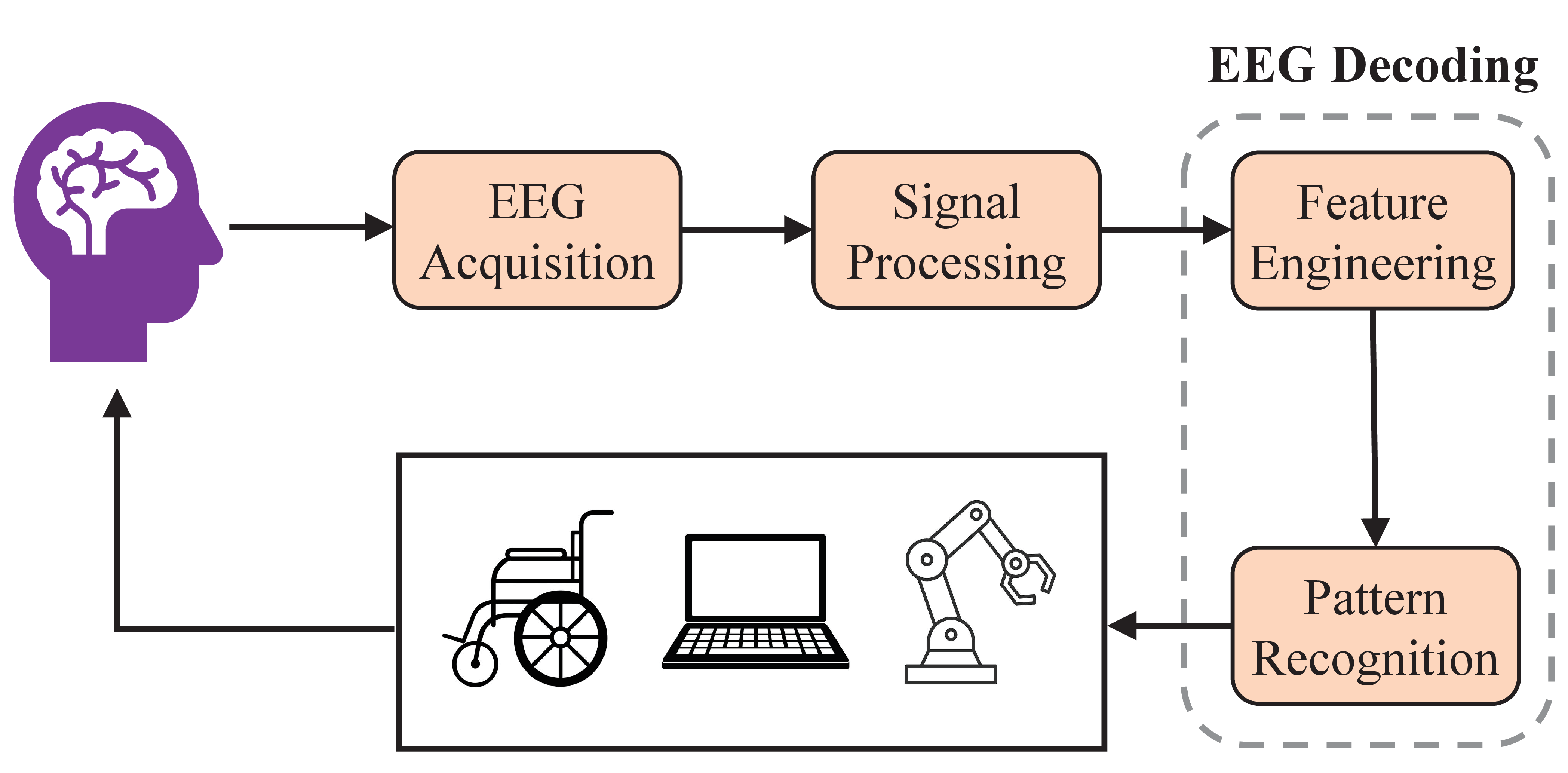}
   \caption{Closed-loop brain-computer interface system.}
   \label{fig:closed_bci}
\end{figure}

Brain-computer interfaces (BCIs) establish direct communication pathways between the human brain and external devices, holding great promise for medical rehabilitation~\cite{yang2022eeg, foong2019assessment}, intelligent control systems~\cite{royer2010eeg}, and sleep stage detection~\cite{altini2021promise}. Figure~\ref{fig:closed_bci} shows the pipeline of closed-loop BCIs. Among various neuroimaging modalities, electroencephalography (EEG) is particularly prominent due to its high temporal resolution, non-invasive acquisition, and cost-effectiveness. Despite these advantages, EEG-based BCIs confront two primary challenges: data scarcity and significant inter-subject variability. These issues considerably hinder the generalization and applicability of EEG decoding models in practical, real-world settings.

Representation learning ~\cite{lian2023watermask, xie2024graph, liu2025umman, liu2024rate} and data generation has emerged as a vital strategy. Representation learning techniques can extract high-order latent features to disentangle underlying neural factors and guide realistic EEG data generation. Conventional data augmentation methods are typically classified into two categories: transformation-based methods and generative model-based methods. Transformation-based approaches employ signal manipulation techniques such as noise addition, amplitude scaling, and frequency shifting. These methods are computationally efficient but often yield limited diversity and physiological realism in the augmented data. Conversely, generative model-based approaches, including generative adversarial networks (GANs)~\cite{altini2021promise}, variational autoencoders (VAEs)~\cite{kingma2013auto}, and diffusion models~\cite{croitoru2023diffusion}, can produce richer and more diverse synthetic data. However, these models typically require large training datasets and significant computational resources, limiting their effectiveness in few-shot scenarios common in BCIs.

Moreover, EEG data inherently possess substantial variability across subjects due to physiological and neuroanatomical discrepancies. Traditional data augmentation methods, predominantly tailored for single-subject contexts, exhibit sharply reduced efficacy when deployed in cross-subject scenarios. Consequently, the development of augmentation strategies specifically designed for cross-subject, few-shot EEG scenarios is critical to advancing practical BCI applications.

\begin{figure}[t]
  \centering
   \includegraphics[width=0.95\linewidth]{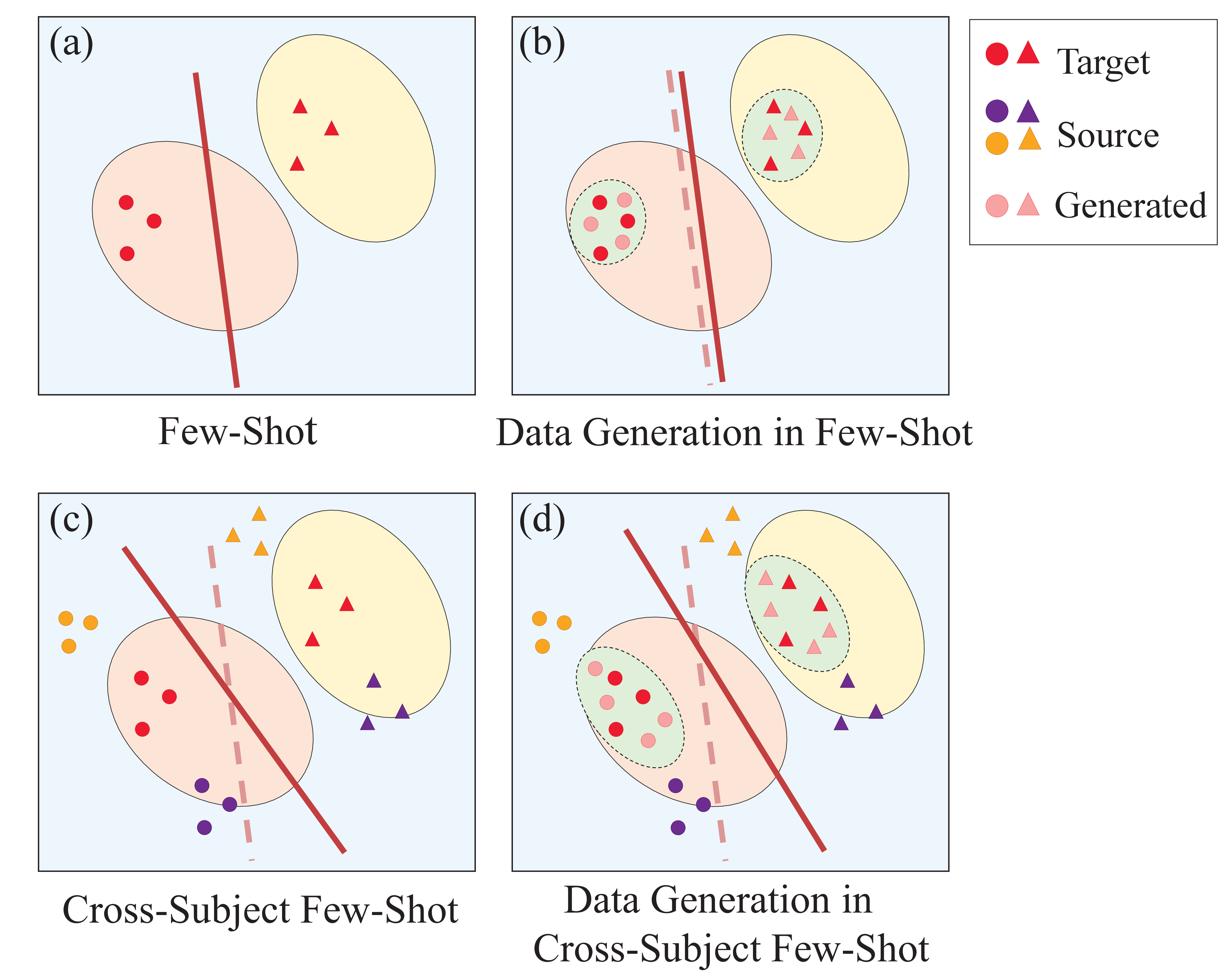}
   \caption{Scenarios of data distributions and generation strategies.}
   \label{fig:scenario}
\end{figure}

As shown in Figure ~\ref{fig:scenario}, in (a), the few‐shot setting with few target samples (red) produces a limited decision boundary. In (b), conventional within-domain augmentation adds synthetic target samples, expanding and smoothing that boundary. In (c), the cross-subject few-shot scenario shows source samples lying outside the target distribution, making transfer ineffective. In (d), data generation approach in cross-subject few-shot scenario, bridging the distribution gap and yielding a refined decision boundary.

In this paper, we propose FusionGen, a novel feature fusion-based EEG data generation framework that efficiently synthesizes diverse EEG signals from scarce samples. By integrating features across samples through a dedicated fusion module and combining them with a lightweight feature extraction and reconstruction pipeline, our architecture ensures both data diversity and trainability under severe data constraints. Comprehensive experiments on multiple publicly available EEG datasets demonstrate that our approach significantly outperforms existing augmentation techniques, yielding notable improvements in classification accuracy.

\begin{enumerate}
\item We propose FusionGen, a few-shot EEG data generation framework that enhances generalization and scalability in brain–computer interface applications.
\item We introduce a feature matching fusion module that integrates cross‐sample features in the latent representation space and reconstructs high-fidelity EEG signals from these fused embeddings.
\item We validate FusionGen on multiple EEG datasets on MI and SSVEP paradigms, showing consistent accuracy improvements in few-shot scenarios.
\end{enumerate}

\section{Related Work}
\label{sec:related}

\subsection{Transformation-Based Approaches}

Wei \textit{et al.}~\cite{wang2018data} proposed noise addition, a simple yet effective method to simulate signal perturbations for EEG augmentation. Xu \textit{et al.}~\cite{zhang2022multi} proposed amplitude scaling, a novel approach that linearly scales EEG signal amplitudes to diversify training data. Wang \textit{et al.}~\cite{wang2024channel} proposed channel reflection, a spatial-domain method exploiting hemispheric symmetry to generate new EEG samples. Wang \textit{et al.}~\cite{wang2025time} proposed DWTaug, a discrete wavelet transform–based augmentation strategy that recombines sub-band signals across samples. Wang \textit{et al.}~\cite{wang2025time} also proposed HHTaug, a Hilbert–Huang transform–based method for cross-sample signal decomposition and reconstruction.

\subsection{Generative Model-Based Methods}

Zhang \textit{et al.}~\cite{zhang2020data} proposed DCGAN, a deep convolutional GAN for synthesizing realistic EEG signals in motor imagery tasks. Luo and Lu~\cite{zhang2021eeg} proposed CWGAN, a conditional Wasserstein GAN to enhance classification accuracy through label-conditioned EEG synthesis. Komolovaite \textit{et al.}~\cite{komolovaite2022deep} performed VAE framework for EEG signal augmentation, learning latent distributions to generate clinically relevant samples. Huang \textit{et al.}~\cite{huang2024eegdfus} adopted diffusion models, modeling EEG denoising via forward-reverse diffusion processes to produce high-fidelity synthetic data.

\subsection{Few-Shot Generative Models}

Wang \textit{et al.}~\cite{ding2022attribute} proposed AGE, an adaptive latent-space feature transformation model for few-shot generative synthesis. Singh \textit{et al.}~\cite{zheng2023my} proposed LSO, a latent space optimization approach modeling category-specific distributions under limited data. Kim \textit{et al.}~\cite{li2023euclidean} proposed HAE, a hyperbolic autoencoder capturing hierarchical semantics to improve few-shot generation diversity. Li \textit{et al.}~\cite{gu2021lofgan} proposed LoFGAN, a local feature fusion GAN that enhances sample diversity by matching and fusing localized features. Chen \textit{et al.}~\cite{zhou2024exact} proposed F2DGAN, a feature-distribution matching GAN combined with variational feature learning for diverse EEG sample synthesis.

In summary, while transformation-based methods are limited by the diversity and quantity of generated data, and generative model-based approaches require large-scale training datasets, few-shot generative models offer a potential solution for EEG augmentation in BCI applications with limited data. Leveraging limited EEG data for generation holds great promise in BCIs.

\section{Method}
\label{sec:method}

\subsection{Problem Definition}

\begin{figure*}[t]
  \centering
   \includegraphics[width=0.95\linewidth]{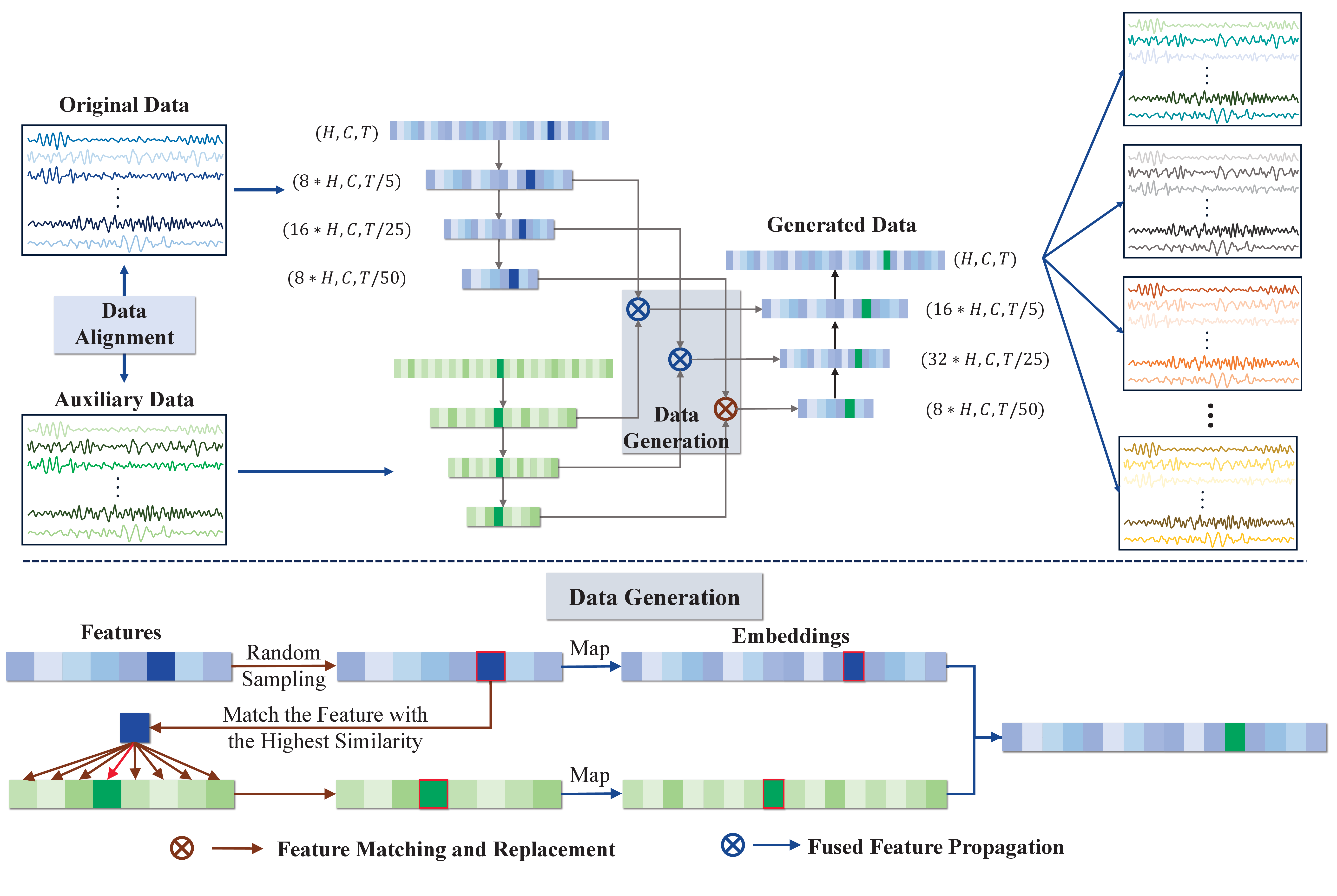}
   \caption{Architecture of proposed FusionGen. Raw and auxiliary trials are first aligned, then encoded into multi-scale features; randomly sampled target features are matched and replaced with source features, propagated through the decoder via skip connections, and finally decoded to produce realistic generated EEG trials.}
   \label{fig:fusiongen}
\end{figure*}

Given $m$ source subjects $S_m$ each providing $n_m$ limited labeled trials $\{X_m^i,y_m^i)\}_{i=1}^{n_m}$, where $X_m^i\in\mathbb{R}^{C\times T}$ and $y_m^i\in{1,\dots,K}$, and a target subject $S_t$ providing $n_t$ labeled trials $\{(X_t^i,y_t^i)\}_{i=1}^{n_t}$ and $n_u$ unlabeled trials $\{X_u^i\}_{i=1}^{n_u}$, where $X_t^i,X_u^i\in\mathbb{R}^{C\times T}$, our goal is to learn a generator $\mathcal{G}$ that synthesizes $n_a$ additional labeled trials $\{(\tilde{X}^{j},y'^{j})\}_{j=1}^{n_a}$ with $\tilde{X}^{j}\in\mathbb{R}^{C\times T}$ and $\tilde{y}^{j}\in{1,\dots,K}$, so that the augmented training set ${(X_m^i,y_m^i)}\cup{(X_t^i,y_t^i)}\cup{(\tilde{X}^{j},\tilde{y}^{j})}$ better approximates the true EEG distribution and improves downstream classification performance on the unlabeled trials ${X_u^i}$. Table~\ref{notations} summarizes the main notations used throughout this paper.

\begin{table}[h]
\centering
\renewcommand{\arraystretch}{1.4}
\caption{Notations used in this paper.}
\label{notations}
\scriptsize
\begin{tabular}{c|>{\centering\arraybackslash}p{0.65\linewidth}}
\toprule
\textbf{Notation} & \textbf{Description} \\
\midrule
$m$ & Number of source subjects. \\
$C$ & Number of EEG channels. \\
$T\;/\;T'$ & Original / compressed time samples. \\
$k$ & Number of replaced features. \\
$S_{m}\;/\;S_{t}$ & Source / target subject. \\
$\{(X_{m}^i,y_{m}^i)\}$\,/\,$\{(X_{t}^i,y_{t}^i)\}$ & Labeled trials of source / target domain. \\
$\bar{X}_{i}$ & Aligned EEG trial. \\
$\tilde{X}_{i}$ & Generated EEG trial. \\
$F_{m}^\ell\;/\;F_{t}^\ell$ & Feature map of source / target at layer $\ell$. \\
$f_{m}^p\;/\;f_{t}^q$ & Source / target latent embeddings. \\
$\hat{F}_{t}$ & Fused feature map after matching and replacement. \\
$\bar{R}$ & Mean covariance matrix of all trials. \\
$(c_{m},x_{m})\;/\;(c_{t},x_{t})$ & Replaced feature indices of source / target. \\
$\mathcal{G}$ & EEG trials generator. \\
$\mathcal{L}_{\mathrm{rec}}$ & Reconstruction loss. \\
\bottomrule
\end{tabular}
\end{table}

\subsection{Input Distribution Alignment}

EEG signals are inherently non-stationary and exhibit substantial variability across sessions and subjects. To mitigate these effects and improve consistency, we adopt Euclidean alignment (EA)~\cite{he2019transfer, wu2025revisiting}, a simple yet effective whitening-based preprocessing step. Given a recording session with $n$ trials $\{X_i\}_{i=1}^n$, where each \(X_i\in\mathbb{R}^{C\times T}\), EA first computes the mean covariance matrix of all trials:
\begin{equation}
\bar{R} \;=\;\frac{1}{n}\sum_{i=1}^n X_i X_i^\top\,. \label{eq:EA-Ref}
\end{equation}
Next, each trial is whitened by
\begin{equation}
\bar{X}_i \;=\;\bar{R}^{-\tfrac{1}{2}}\,X_i\,. \label{eq:s-EA}
\end{equation}
Since 
\begin{equation}
\frac{1}{n}\sum_{i=1}^n \bar{X}_i\,\bar{X}_i^\top
=\bar{R}^{-\tfrac{1}{2}}
\Bigl(\tfrac{1}{n}\sum_{i=1}^nX_iX_i^\top\Bigr)
\bar{R}^{-\tfrac{1}{2}}
=I,    \label{eq:identity}
\end{equation}
the second-order statistics of the aligned trials become identity, effectively reducing covariance shifts. The aligned set \(\{\bar{X}_i\}_{i=1}^n\) then replaces the raw trials in all subsequent processing steps. By performing input distribution alignment, we substantially suppress marginal distribution differences among subjects, facilitating downstream feature extraction and classification.  
\subsection{Feature Matching Fusion}

As shown in Figure ~\ref{fig:fusiongen}, we employ a U-Net-shaped encoder–decoder network and perform pairwise feature fusion between source and target feature representations in the latent space at each latent space, injecting the fused features into the model via skip connections. Specifically, after EA and initial feature extraction, we obtain source and target feature maps:
\begin{equation}
F_m,\,F_t\;\in\;\mathbb{R}^{C\times T'}.
\end{equation}

To inject cross‐sample diversity while preserving class semantics, we perform feature matching fusion in four steps:

1. \textbf{Vectorization}: Reshape each map into $N=C\times T'$ column vectors:  
\begin{equation}
   \{f_m^p\in\mathbb{R}^d\}_{p=1}^N,\quad \{f_t^q\in\mathbb{R}^d\}_{q=1}^N,
\end{equation}
   where $d$ is the channel depth.

2. \textbf{Index Selection}:  
   Randomly sample $k$ target positions:
   \begin{equation}
   \mathcal{Q}=\{q_1,\dots,q_k\}\subset\{1,\dots,N\}.
   \end{equation}

3. \textbf{Cosine Matching}:  
   For each $q\in\mathcal{Q}$ compute similarities to all source vectors:
   \begin{align}
   s_{pq} &= \frac{\langle f_t^q,\,f_m^p\rangle}{\|f_t^q\|\;\|f_m^p\|}, \label{eq:cosine-sim-detailed} \\
   p^*_q  &= \arg\max_{p}\,s_{pq}.          \label{eq:best-match-detailed}
   \end{align}
   
   We then substitute each target representation embeddings with its most similar source feature:
   \begin{equation}
   f_t^q \;\leftarrow\; f_m^{p^*_q}\,,\quad \forall\,q\in\mathcal{Q}.
   \end{equation}

4. \textbf{Map Reconstruction}:  
   Reassemble the modified set of vectors into the fused feature map:
   \begin{equation}
     \hat F_t \;=\;\mathrm{reshape}\bigl(\{f_t^q\}_{q=1}^N\bigr)
     \;\in\;\mathbb{R}^{C\times T'}.
   \end{equation}
   
This procedure injects rich, cross‐subject patterns into the target features without discarding their original label‐specific information.

To ensure the fused features permeate every decoder skip‐connection, we propagate them hierarchically across all $L$ encoder–decoder layers. Denote the cumulative downsampling factor at layer $\ell$ by $r_\ell$. A base‐layer coordinate $(c,t)$ then corresponds to a region in layer $\ell$:
\begin{align}
R_\ell(c,t)
&=\bigl[\lfloor c/r_\ell\rfloor : \lceil c/r_\ell\rceil\bigr]
 \times
 \bigl[\lfloor t/r_\ell\rfloor : \lceil t/r_\ell\rceil\bigr].
\label{eq:prop-window}
\end{align}
Within each region $R_\ell(c,t)$ of the $\ell$‐th target feature map $F_t^\ell$, we perform:
\begin{align}
F_t^\ell(u,v)
&\leftarrow 
F_m^\ell\bigl(\lfloor u\cdot r_\ell\rfloor,\lfloor v\cdot r_\ell\rfloor\bigr),
\quad
(u,v)\in R_\ell(c,t).
\label{eq:prop-replacement-detailed}
\end{align}

By applying Eqs.~(\ref{eq:prop-window}) and (\ref{eq:prop-replacement-detailed}) for all selected $(c,t)$ across $\ell=1,\dots,L$, every decoder skip‐connection integrates the fused features, preserving semantic alignment and enabling high‐fidelity reconstruction under scarce data conditions.

\subsection{Network Architecture and Training}

\begin{figure}[t]
  \centering
   \includegraphics[width=0.95\linewidth]{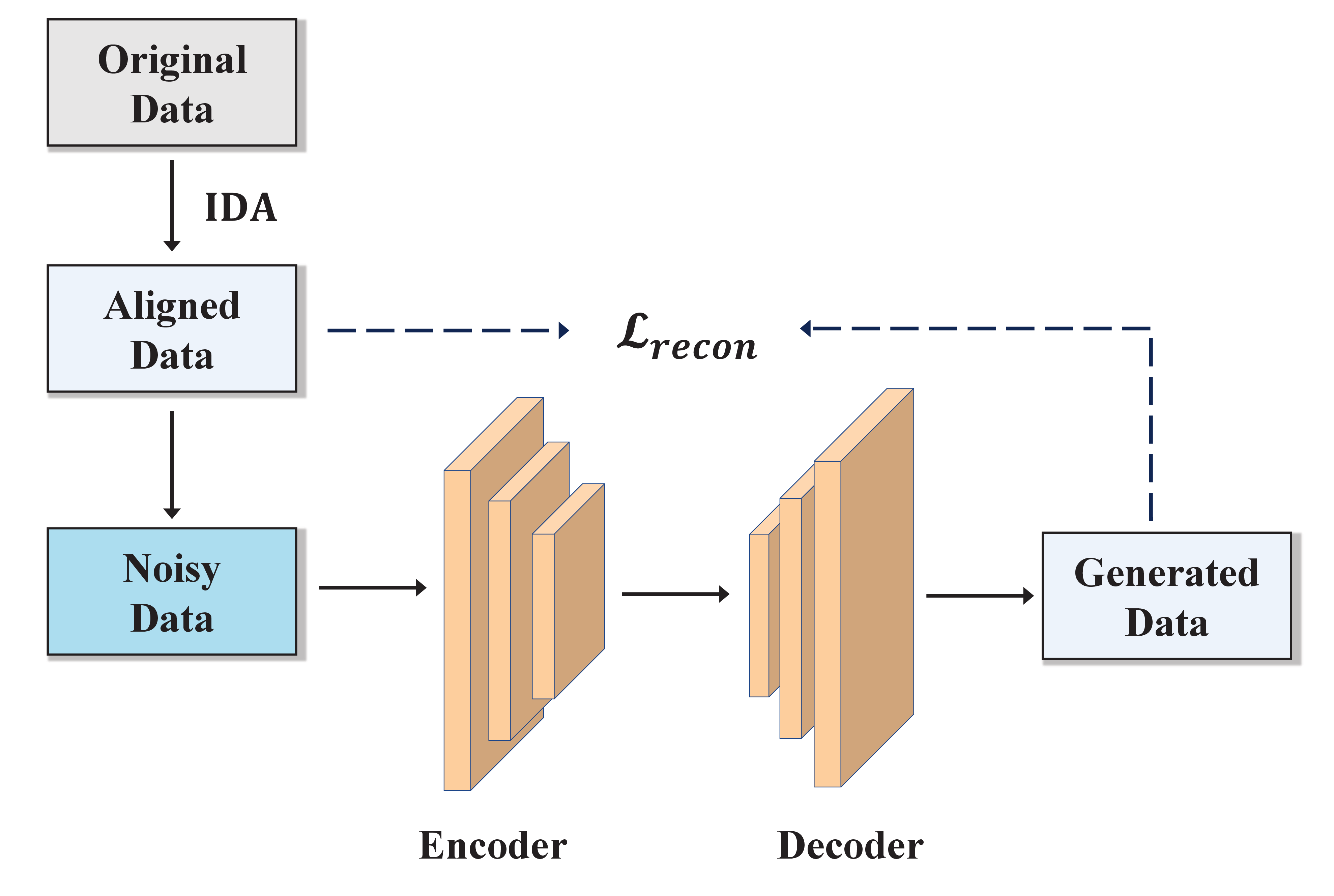}
   \caption{Overview of the proposed EEG data generation pipeline.}
   \label{fig:recon}
\end{figure}

Figure~\ref{fig:recon} illustrates the overall generator architecture, which follows a U-Net–shaped encoder–decoder design with integrated fusion modules.  Given an aligned input trial $\bar{X}\in\mathbb{R}^{C\times T}$, the encoder comprises three convolutional blocks that reduce the temporal dimension by factors $r_1=5$, $r_2=5$, and $r_3=2$. Each bottleneck feature map $F_t\in\mathbb{R}^{C\times T'}$ is then passed through our Feature Matching Fusion modules to produce a fused map $\hat F_t$.

The decoder mirrors the encoder with three transposed convolutions that upsample by $(r_3,r_2,r_1)$ and channel depths reversed. A final transposed convolution restores the output $\tilde X\in\mathbb{R}^{C\times T}$.

To train the network, we adopt a denoising autoencoder strategy: each aligned trial $X$ is perturbed with additive Gaussian noise to yield $\tilde X$, and the model is optimized to reconstruct the clean signal.  We minimize the mean squared error over the aligned dataset:
\begin{align}
\mathcal{L}_{\mathrm{recon}}
&= \frac{1}{n}\sum_{i=1}^n \bigl\|\bar{X}_i - \tilde X_i\bigr\|_2^2.
\end{align}

After convergence, the trained generator $\mathcal{G}$ synthesizes augmented trials by applying the fusion modules to new inputs, enabling diverse and physiologically plausible EEG data generation in few-shot scenarios.

\section{Experiment}
\label{sec:exp}

\subsection{Datasets}

We consider the motor imagery (MI) and steady-state visual evoked potentials (SSVEP) paradigm in our experiment, the most widely adopted protocol in BCI research. MI~\cite{pfurtscheller2001motor} is the cognitive process of imagining the movement of different body parts without actually moving them. Three MI datasets and one SSVEP dataset, all from the mother of all BCI benchmark (MOABB)~\cite{Jayaram2018} and summarized in Table~\ref{tab:datasets}, were utilized in the experiments.

\begin{table*}[htpb]
\centering
\renewcommand{\arraystretch}{1.2}
\caption{Summary of the three MI datasets.}
\label{tab:datasets}
\resizebox{\textwidth}{!}{%
\begin{tabular}{c|c|c|c|c|c|c|c}
\toprule
BCI & \multirow{2}{*}{Dataset} & Number of & Number of & Sampling & Trial Length & Number of   & \multirow{2}{*}{Class Labels} \\
Paradigm &                         & Subjects  & Channels  & Rate (Hz) & (seconds)    & Total Trials &                       \\
\midrule
\multirow{3}{*}{MI}
 & BNCI2014001 & 9  & 22 & 250 & 4 & 2592 & left \& right hand, tongue, feet \\
 & BNCI2014002 & 14 & 15 & 512 & 5 & 1400 & left hand, right hand            \\
 & Zhou2016    & 4  & 14 & 250 & 5 & 1842 & left hand, right hand, feet      \\

 \midrule
\multirow{1}{*}{SSVEP}
 & Nakanishi2015 & 9  & 8 & 256 & 4.15 & 1620 & 12 different stimuli \\
 
\bottomrule
\end{tabular}%
}
\end{table*}

The three MI datasets used in this study are described below:
\begin{enumerate}
\item BNCI2014001~\cite{tangermann2012review}: BNCI2014001 dataset contains EEG data from 9 subjects performing four MI tasks: left hand, right hand, both feet, and tongue. Each subject participated in two sessions, with each session consisting of 6 runs, yielding a total of 288 trials per session.
\item BNCI2014002~\cite{steyrl2016random}: BNCI2014002 dataset includes EEG data from 13 participants performing sustained MI of the right hand and feet. The session consists of eight runs, with 50 trials per class for training and 30 trials for validation. EEG was recorded at 512 Hz from 15 electrodes, including C3, Cz, and C4, with a biosignal amplifier and active Ag/AgCl electrodes.
\item Zhou2016 ~\cite{zhou2016fully}: Zhou2016 dataset includes EEG data from 4 subjects performing three MI tasks: left hand, right hand, and feet. Each subject participated in three sessions, with each session consisting of two runs of 75 trials (25 trials per class).
\end{enumerate}

The SSVEP-based Nakanishi2015 dataset used in this study are described below:

\begin{enumerate}
\item Nakanishi2015~\cite{nakanishi2015comparison}: Nakanishi2015 dataset is an SSVEP‐based EEG benchmark comprising recordings from 9 subjects with 8 channels across 12 visual stimulation classes (frequencies ranging from 9.25 Hz to 14.75 Hz in 0.5 Hz steps), each with 15 trials of 4.15 s duration sampled at 256 Hz in a single session.
\end{enumerate}

\subsection{Settings}

All data were preprocessed using the standard MOABB pipeline. Specifically, EEG recordings were downsampled to $250\,\mathrm{Hz}$, bandpass‐filtered to $8$–$32\,\mathrm{Hz}$, and each trial was truncated to the first $4\,\mathrm{s}$ (1000 samples).

To prevent data leakage, we split training and test sets in temporal order rather than at random.  In the cross‐subject evaluation, we adopt a leave‐one‐subject‐out (LOSO) protocol: for each target subject, the first $n_{\mathrm{train}}$ continuous trials per class are used for augmentation and model training, and the remaining trials for testing.  In the within‐subject scenario, we similarly split each subject’s trials by selecting the first $n_{\mathrm{train}}$ trials per class for training and the remainder for testing.

The generator network is trained as a denoising autoencoder using the Adam optimizer with learning rate $0.01$, batch size $64$.  Feature matching fusion selects $k=0.2\times N$ vectors (where $N=C\times T'$) and uses a noise intensity coefficient of $5$.  To mitigate randomness from the sampling and fusion steps, each experiment is repeated $10$ times and results are averaged.

For downstream classification, we apply the classic CSP-LDA pipeline~\cite{blankertz2007optimizing}, extracting $10$ spatial filters via common spatial patterns (CSP) and classifying with a linear discriminant analysis (LDA) model.  

\subsection{Main Results}

\begin{figure}[t]
  \centering
   \includegraphics[width=0.95\linewidth]{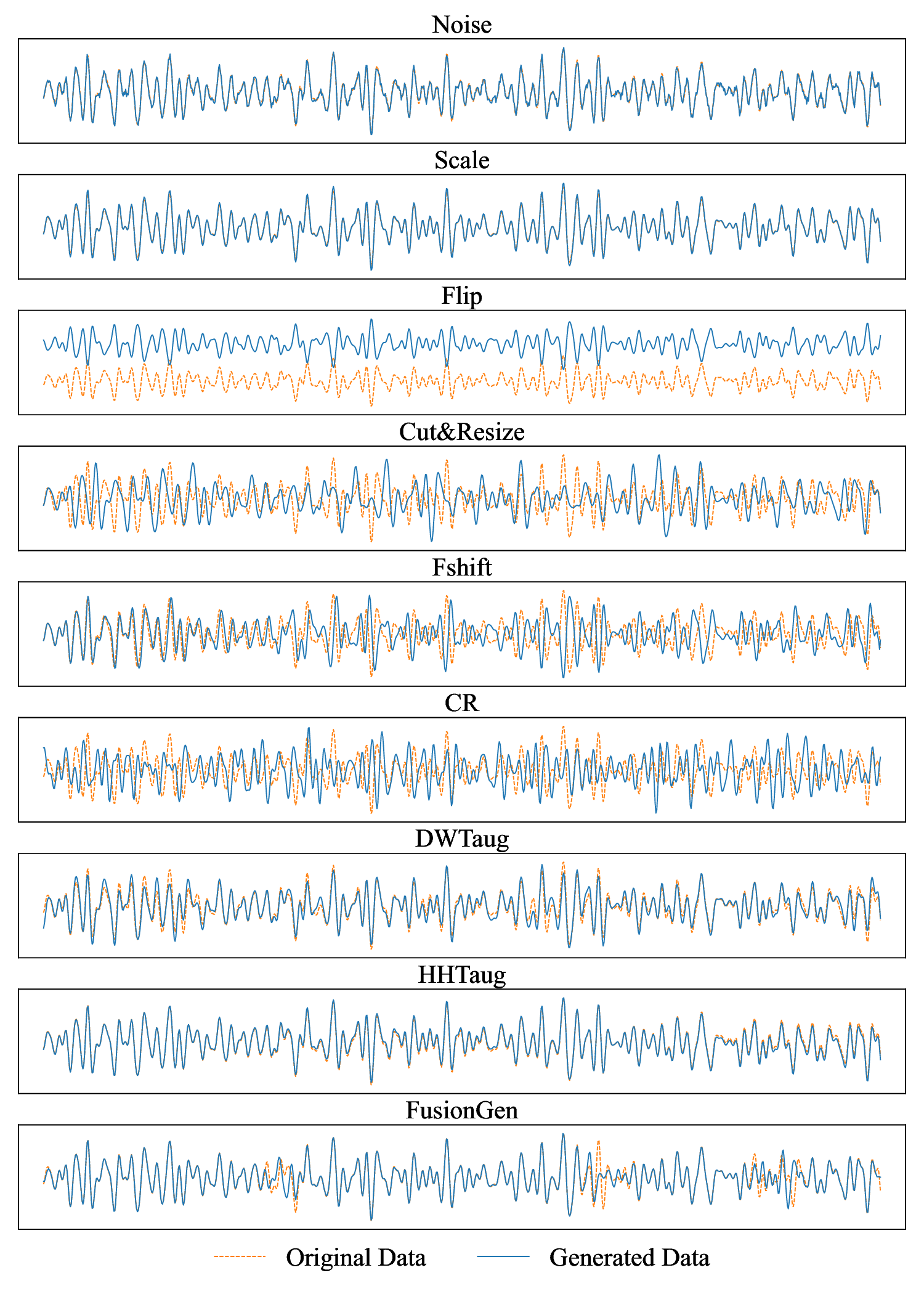}
   \caption{Visualization of generated EEG signals by various augmentation methods.}
   \label{fig:vis_gen}
\end{figure}

Figure~\ref{fig:vis_gen} visualizes the comparison between original and generated EEG signals by various augmentation methods. Table~\ref{tab:results_cross} reports cross‐subject few‐shot accuracy for three MI datasets under varying numbers of calibration trials.  FusionGen consistently outperforms both no‐augmentation and most baselines. On BNCI2014001, FusionGen achieves 57.78\% (vs.\ 53.62\% without augmentation), a 4.16\% enhancement; on BNCI2014002 it reaches 73.97\% (vs.\ 72.46\%), a 1.51\% boost; and on Zhou2016 it attains 65.84\% (vs.\ 62.04\%), a 3.80\% improvement.  

Table~\ref{tab:results_within} shows within‐subject few‐shot accuracy.  On BNCI2014001, FusionGen achieves 55.86\% (vs.\ 52.53\% without augmentation), a 3.33\% enhancement; on BNCI2014002 it reaches 70.47\% (vs.\ 65.83\%), a 4.64\% boost; and on Zhou2016 it attains 62.53\% (vs.\ 54.63\%), a 7.90\% improvement.

In most few-shot settings (7, 10, 15), FusionGen achieves good performance, validating its ability to bridge source and target distributions under extreme data scarcity.

Furthermore, we also validate the performance of FusionGen on the SSVEP‐based Nakanishi2015 dataset, as reported in Table~\ref{tab:results_ssvep}, demonstrating consistent improvements across different BCI paradigms.

\begin{table*}[htpb]
\centering
\renewcommand{\arraystretch}{1.2}
\caption{Classification accuracy (\%) of various data augmentation approaches on three MI datasets under a cross-subject few-shot evaluation.}
\label{tab:results_cross}
\small
\begin{tabular}{w{c}{1.7cm}|w{c}{0.7cm}|*{10}{w{c}{1cm}}}
\toprule
Dataset & Trials & None & Noise & Scale & Flip & Cut\&Resize & Fshift & CR & DWTaug & HHTaug & FusionGen \\
\midrule
\multirow{4}{*}{BNCI2014001}
 & 7    & 52.37 & 52.26 & 51.19 & 52.13 & 52.96 & 52.15 & 53.16 & \textbf{54.97} & 54.17 & \underline{54.85} \\
 & 10   & 53.56 & 53.29 & 52.73 & 53.38 & 52.84 & 53.14 & 54.64 & 57.00 & \underline{57.43} & \textbf{58.02} \\
 & 15   & 54.93 & 53.41 & 53.65 & 54.26 & 54.95 & 54.88 & 55.18 & 57.10 & \underline{57.16} & \textbf{60.48} \\
 & Avg. & 53.62 & 52.99 & 52.52 & 53.26 & 53.58 & 53.39 & 54.33 & \underline{56.36} & 56.25 & \textbf{57.78} \\
\midrule
\multirow{4}{*}{BNCI2014002}
 & 7    & 71.61 & 72.56 & 71.25 & 70.89 & 72.26 & 71.73 & 67.08 & \underline{72.62} & 71.79 & \textbf{73.56} \\
 & 10   & 72.80 & 73.27 & 73.27 & 68.75 & 72.20 & 72.80 & 70.36 & \textbf{73.75} & \underline{73.57} & 73.45 \\
 & 15   & 72.98 & 72.44 & 73.63 & 69.82 & 72.80 & 73.69 & 71.25 & 74.17 & \underline{74.64} & \textbf{74.89} \\
 & Avg. & 72.46 & 72.76 & 72.72 & 69.82 & 72.42 & 72.74 & 69.56 & \underline{73.51} & 73.33 & \textbf{73.97} \\
\midrule
\multirow{4}{*}{Zhou2016}
 & 7    & 62.31 & 62.46 & 63.50 & 57.38 & 58.74 & 61.76 & 63.34 & \underline{63.97} & \textbf{64.00} & 63.71 \\
 & 10   & 59.56 & 57.92 & 60.84 & 56.69 & 58.30 & 64.52 & 64.43 & 64.62 & \textbf{68.88} & \underline{64.95} \\
 & 15   & 64.26 & 64.14 & 66.45 & 68.78 & 64.77 & 64.84 & 68.40 & 68.50 & \textbf{69.72} & \underline{68.85} \\
 & Avg. & 62.04 & 61.51 & 63.60 & 60.95 & 60.60 & 63.71 & 65.39 & 65.70 & \textbf{67.53} & \underline{65.84} \\
\bottomrule
\end{tabular}
\end{table*}

\begin{table*}[htpb]
\centering
\renewcommand{\arraystretch}{1.2}
\caption{Classification accuracy (\%) of various data augmentation approaches on three MI datasets under a within-subject few-shot evaluation.}
\label{tab:results_within}
\small
\begin{tabular}{w{c}{1.7cm}|w{c}{0.7cm}|*{10}{w{c}{1cm}}}
\toprule
Dataset & Trials & None & Noise & Scale & Flip & Cut\&Resize & Fshift & CR & DWTaug & HHTaug & FusionGen \\
\midrule
\multirow{4}{*}{BNCI2014001}
 & 10  & 47.56 & 49.82 & 48.56 & 49.80 & 48.39 & 46.88 & \textbf{52.55} & 47.71 & 44.41 & \underline{51.93} \\
 & 14  & 53.55 & 54.01 & 53.60 & \underline{55.32} & 54.76 & 54.33 & 53.43 & 53.88 & 45.39 & \textbf{56.83} \\
 & 20  & 56.48 & \underline{57.08} & 56.73 & 56.53 & 54.79 & 56.15 & 55.09 & 56.20 & 45.94 & \textbf{58.81} \\
 & Avg.& 52.53 & 53.64 & 52.96 & \underline{53.88} & 52.65 & 52.45 & 53.69 & 52.60 & 45.25 & \textbf{55.86} \\
\midrule
\multirow{4}{*}{BNCI2014002}
 & 10  & 59.74 & 65.13 & 62.37 & 66.09 & 66.35 & 62.44 & 61.47 & 60.45 & \underline{67.95} & \textbf{68.15} \\
 & 14  & 66.67 & 69.49 & 67.50 & 69.29 & 67.88 & 66.99 & 68.59 & 66.35 & \textbf{71.54} & \underline{71.21} \\
 & 20  & 71.09 & 69.87 & 70.13 & 70.90 & 71.22 & 70.71 & 68.53 & 71.22 & \underline{71.86} & \textbf{72.04} \\
 & Avg.& 65.83 & 68.16 & 66.67 & 68.76 & 68.48 & 66.71 & 66.20 & 66.01 & \underline{70.45} & \textbf{70.47} \\
\midrule
\multirow{4}{*}{Zhou2016}
 & 10  & 47.98 & 53.49 & 54.98 & 52.46 & 50.44 & 49.01 & \textbf{63.59} & 49.12 & \underline{61.61} & 58.31 \\
 & 14  & 52.98 & 55.75 & 53.85 & 57.86 & 55.86 & 55.30 & \textbf{65.55} & 52.44 & \underline{65.31} & 63.00 \\
 & 20  & 62.92 & 65.62 & 68.15 & 67.25 & 67.37 & 63.39 & \textbf{71.11} & 62.00 & \underline{68.27} & 66.27 \\
 & Avg.& 54.63 & 58.29 & 58.99 & 59.19 & 57.89 & 55.90 & \textbf{66.75} & 54.52 & \underline{65.06} & 62.53 \\
\bottomrule
\end{tabular}
\end{table*}

\begin{table*}[htpb]
\centering
\renewcommand{\arraystretch}{1.2}
\caption{Classification accuracy (\%) of various data augmentation approaches on Nakanishi2015 datasets under a cross-subject 1-shot evaluation.}
\label{tab:results_ssvep}
\small
\begin{tabular}{w{c}{1.7cm}|w{c}{1.7cm}|*{10}{w{c}{0.9cm}}}
\toprule
Paradigm & Approaches & S1 & S2 & S3 & S4 & S5 & S6 & S7 & S8 & S9 & Avg. \\
\midrule
\multirow{9}{*}{SSVEP}
& None & 53.65 & 34.11 & 77.08 & 80.73 & 83.85 & 84.90 & 79.43 & 71.09 & 86.20 & 72.34 \\
& Noise & 64.84 & \underline{52.08} & 88.28 & 84.64 & \textbf{96.09} & 94.27 & 85.68 & 84.64 & \underline{95.31} & 82.87 \\
& Scale & \underline{69.01} & 47.92 & 90.36 & 89.58 & 89.06 & \textbf{99.22} & \textbf{91.41} & 85.68 & 93.23 & 83.94 \\
& Flip & 40.62 & 18.23 & 45.57 & 36.72 & 71.61 & 39.06 & 59.11 & 39.06 & 42.19 & 43.58 \\
& Cut\&Resize & 49.22 & 26.04 & 67.45 & 61.20 & 76.30 & 75.52 & 73.18 & 56.77 & 72.66 & 62.04 \\
& Fshift & 57.55 & 45.83 & 91.41 & 89.84 & 91.93 & 96.09 & 87.24 & \underline{86.46} & 87.76 & 81.57 \\
& DWTaug & 66.67 & 48.70 & \textbf{95.31} & \underline{95.31} & \underline{94.01} & 93.23 & \underline{90.36} & 82.03 & 93.49 & \underline{84.35} \\
& HHTaug & 66.93 & 47.40 & \underline{92.97} & 81.77 & 91.93 & \underline{96.88} & 85.94 & 81.77 & 91.15 & 81.86 \\
& FusionGen & \textbf{76.56} & \textbf{69.01} & 82.29 & \textbf{97.66} & 72.92 & \underline{96.88} & 80.21 & \textbf{86.72} & \textbf{100.0} & \textbf{84.69} \\

\bottomrule
\end{tabular}
\end{table*}

\subsection{Effectiveness of FusionGen Integration}

\begin{figure}[t]
  \centering
   \includegraphics[width=0.95\linewidth]{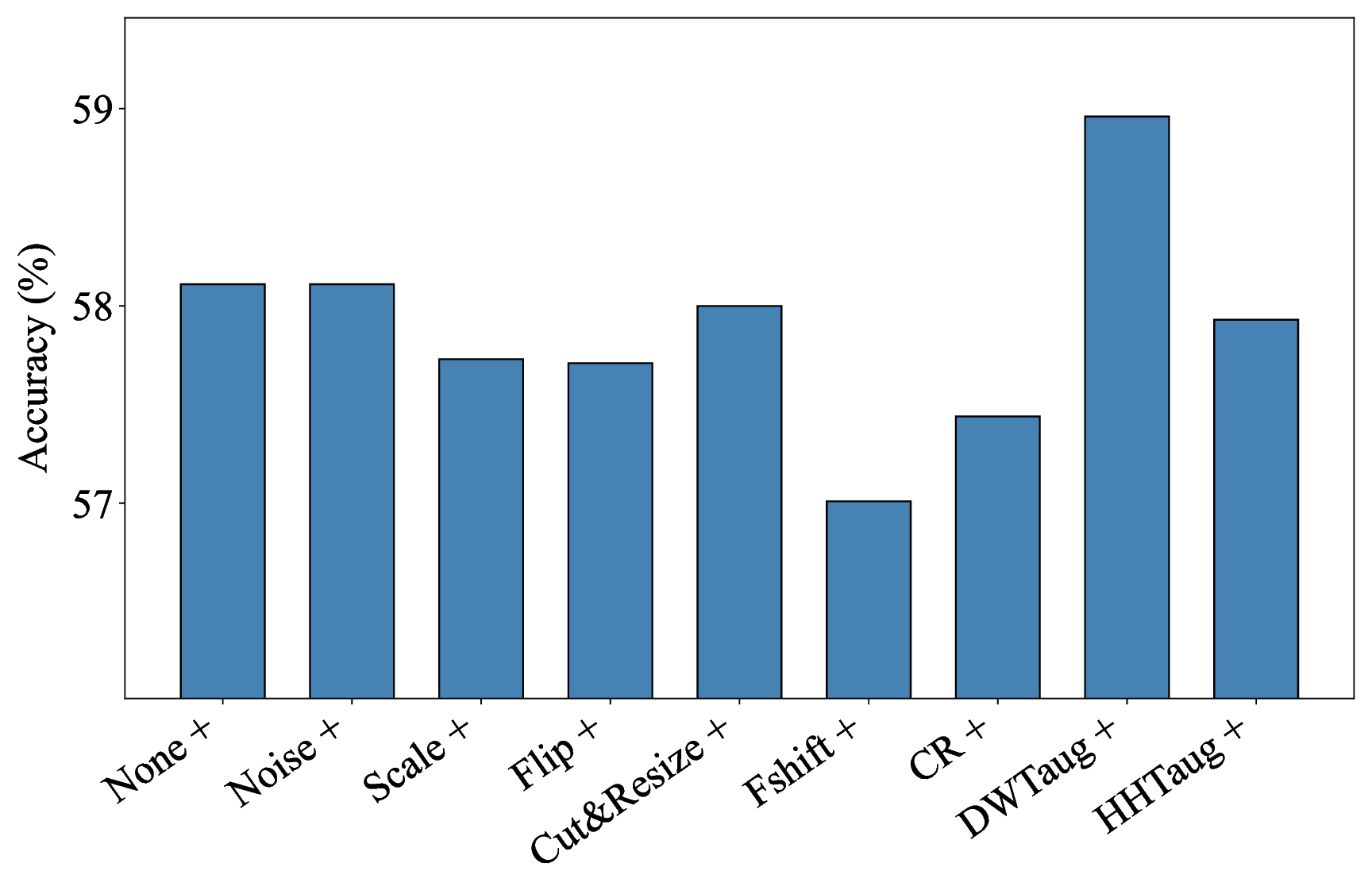}
   \caption{Accuracy of baseline augmentations combined with FusionGen on BNCI2014001 with 10 calibration trials.}
   \label{fig:fusion_integration}
\end{figure}

To assess the compatibility of FusionGen with standard augmentation techniques, we appended our feature‐fusion generator to each baseline approach and evaluated the combined performance on BNCI2014001 with 10 calibration trials. As shown in Figure~\ref{fig:fusion_integration}, all FusionGen augmented pipelines outperform their standalone counterparts. Notably, combining FusionGen with DWTaug yields the highest accuracy (58.96\%), representing a further enhancement over DWTaug alone. Even methods that individually provided modest improvements, such as Noise or Scale—benefit from integration with FusionGen, demonstrating consistent boosts of 0.5-1.0\%. These results confirm that FusionGen can be seamlessly integrated with diverse augmentation strategies to deliver complementary enhancements, underscoring its general applicability in EEG‐based BCI.  

\subsection{Analysis of Generated Sample Distribution}

\begin{figure}[htbp]
\centering
\subfloat[\label{fig:t_sne-a}]{\includegraphics[width=0.428\linewidth]{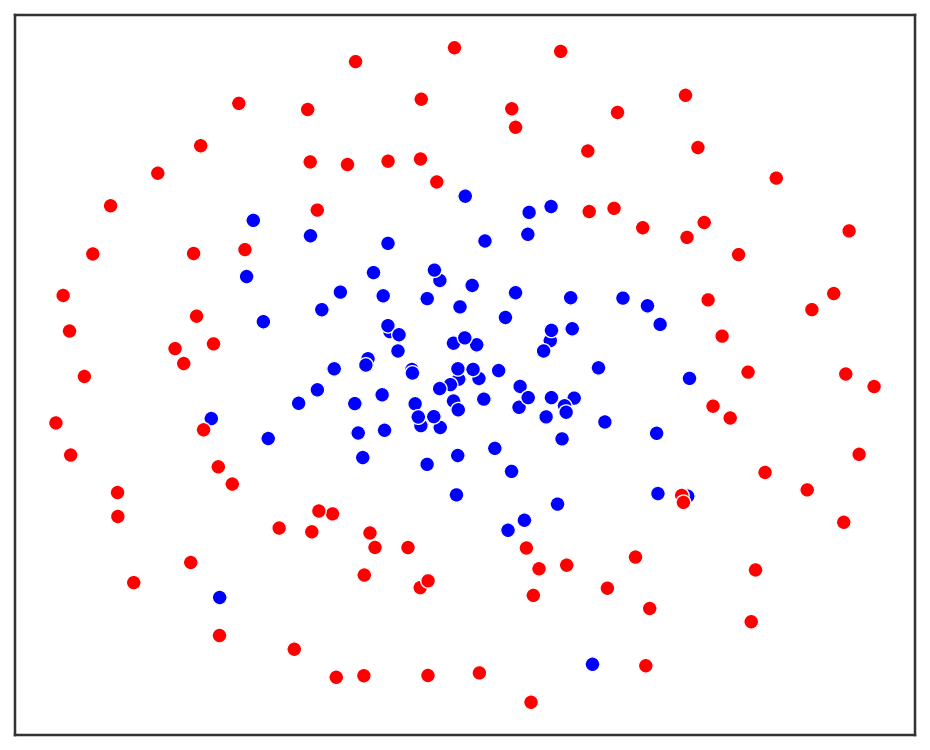}}
\hfill
\subfloat[\label{fig:t_sne-b}]{\includegraphics[width=0.56\linewidth]{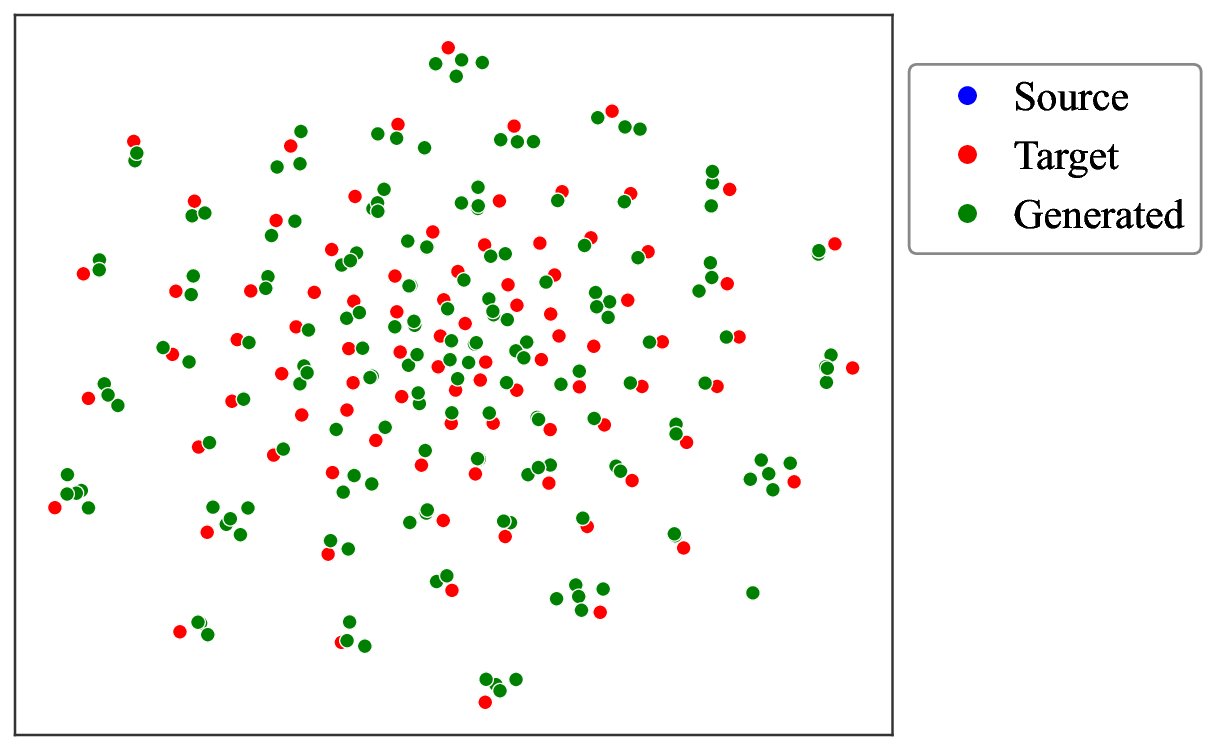}}
\caption{$t$-SNE visualization of the latent distributions on BNCI2014001. (a) Source data and target data. (b) Generated data and target data.}
\label{fig:t_sne}
\end{figure}

Different subjects exhibit substantial EEG variability, creating a large gap between source and target domains. In practical BCI applications, we aim to minimize, or even eliminate the calibration, resulting in very few target trials. As Figure~\ref{fig:t_sne} shows, FusionGen generates abundant synthetic trials (green) that closely follow the true target distribution (red), effectively filling the sparse region around the scarce target samples. This demonstrates that FusionGen can faithfully mimic subject-specific EEG characteristics and achieve high‐quality data generation under extreme few‐shot conditions.  

\section{Discussion}
\label{sec:discussion}

\subsection{Applications}

In transfer‐learning scenarios for BCIs, reducing or eliminating per‐subject calibration is highly desirable. However, most existing methods require target domain trials to align or adapt source models, limiting their practicality when only limited calibration trials are available ~\cite{liu2025spatial, zheng2024semi}.  FusionGen addresses this bottleneck by synthesizing large volumes of target‐like EEG data from minimal samples, enabling robust source–target alignment and downstream model adaptation with few calibration.

Furthermore, EEG data collection remains costly and time‐consuming, and publicly available datasets often lack the scale needed for training large models. While researchers are progressing toward foundation models for BCIs~\cite{liu2025mirepnet, jiang2024large, wang2025eegpt}, the scarcity of large-scale, high-quality datasets remains a fundamental constraint. CLEAN-MI ~\cite{liu2025clean} attempts to filter out high-quality motor imagery data, but is still limited by insufficient data. This paper proposed FusionGen, providing a scalable solution by generating physiologically plausible EEG signals that preserve the statistical properties of real data distributions. This synthetic data generation capability holds significant potential for advancing BCI research, as it can supply unlimited, diverse, and distribution-matched training samples, thereby facilitating the development of more robust and generalizable foundation models.

\subsection{Hyperparameter Analysis}

\begin{figure}[t]
  \centering
   \includegraphics[width=0.95\linewidth]{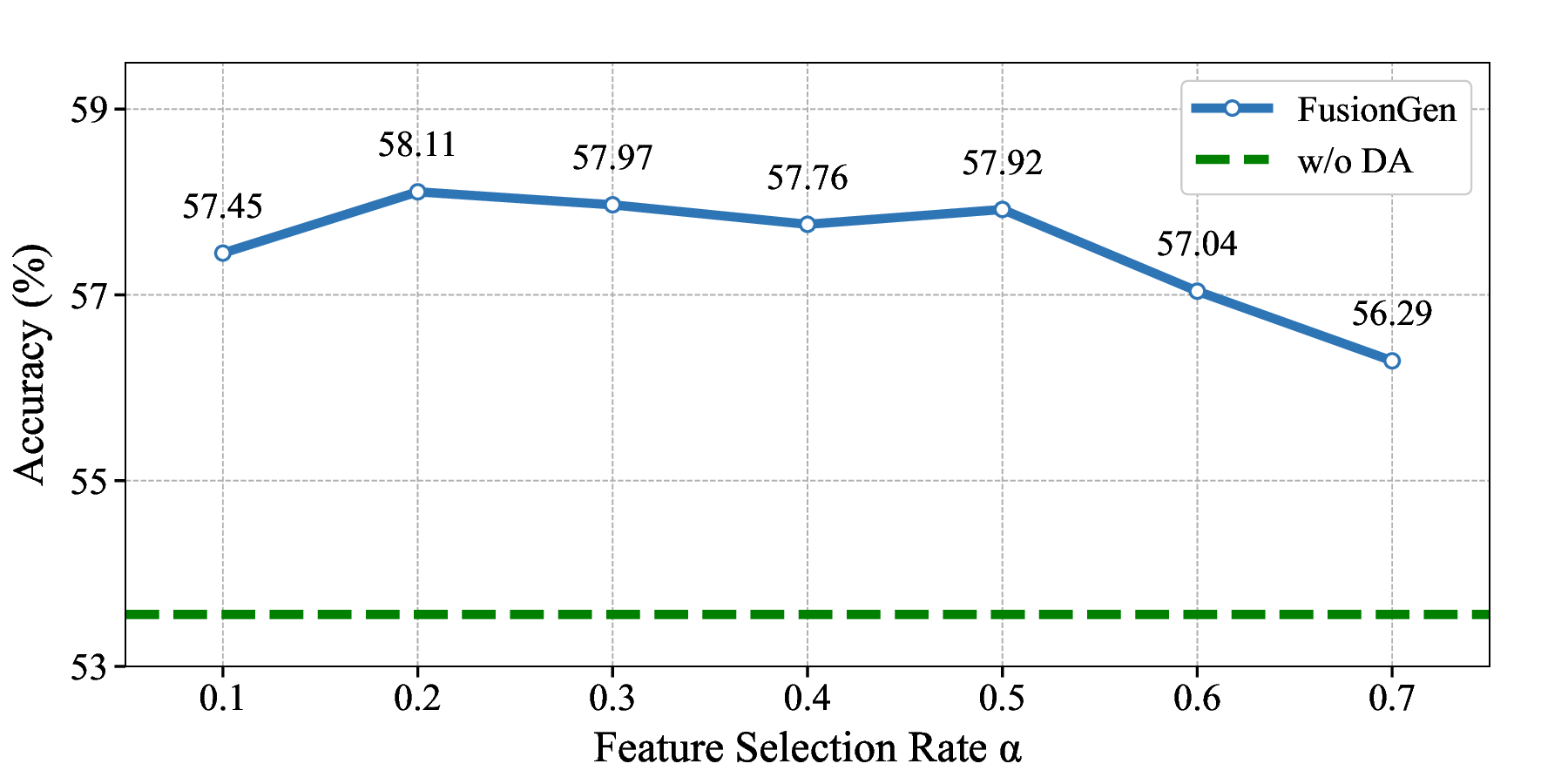}
   \caption{Sensitivity of classification accuracy to feature selection rate $\alpha$}
   \label{fig:sensitivity}
\end{figure}

We analyze the impact of the feature selection rate $\alpha$ on classification accuracy (Figure~\ref{fig:sensitivity}). As $\alpha$ varies from 0.1 to 0.5, accuracy remains stable around 57.7-58.1\%, indicating that FusionGen is robust to the ratio of replaced features. However, when $\alpha$ exceeds 0.6, performance begins to decline (57.0\% at 0.6, 56.3\% at 0.7), suggesting that overly aggressive feature replacement can degrade the representation learning capacity of the latent space.

\begin{figure}[htbp]
\centering
\subfloat[\label{fig:spectrum_ratio-a}]{\includegraphics[width=0.50\linewidth]{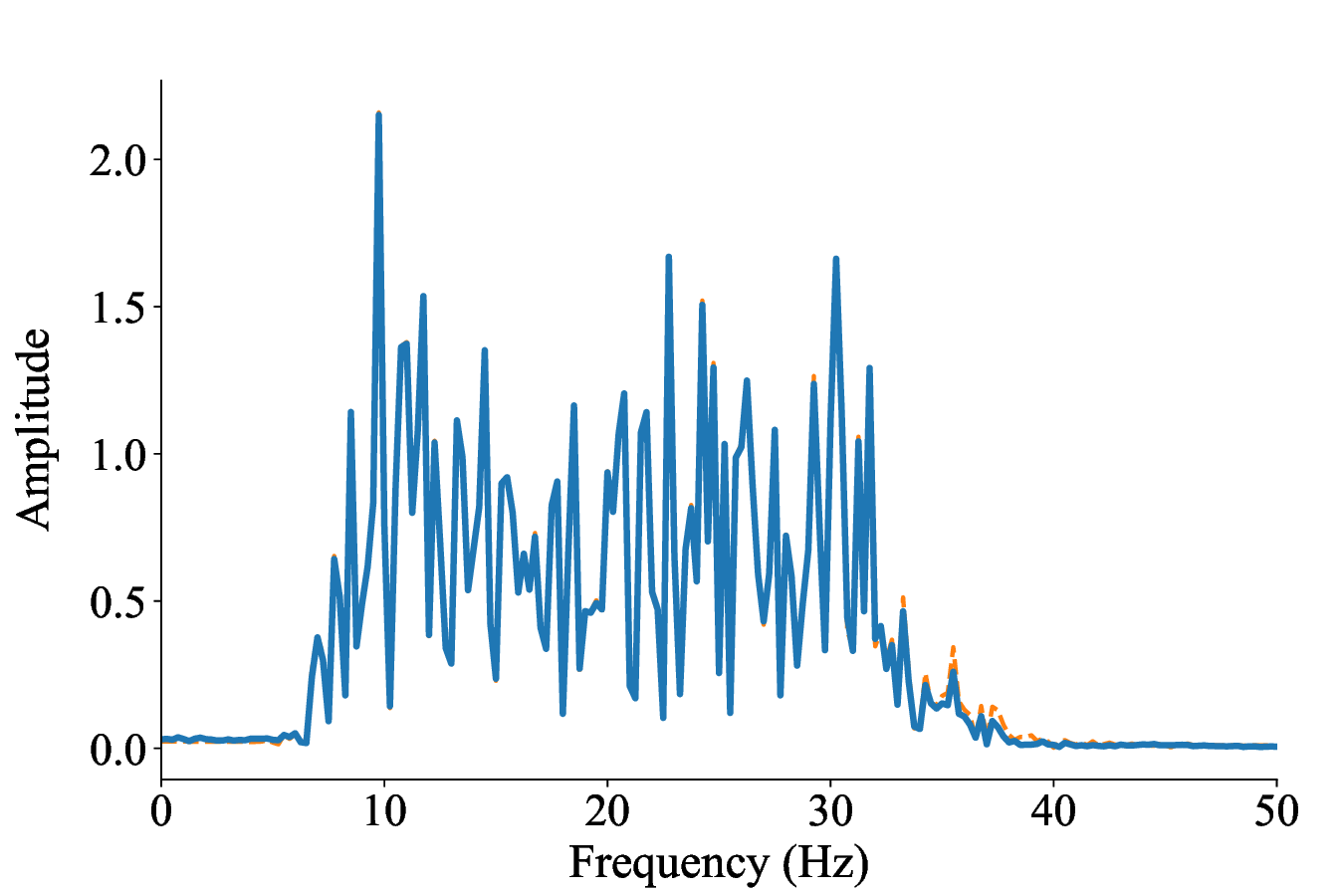}}
\hfill
\subfloat[\label{fig:spectrum_ratio-b}]{\includegraphics[width=0.50\linewidth]{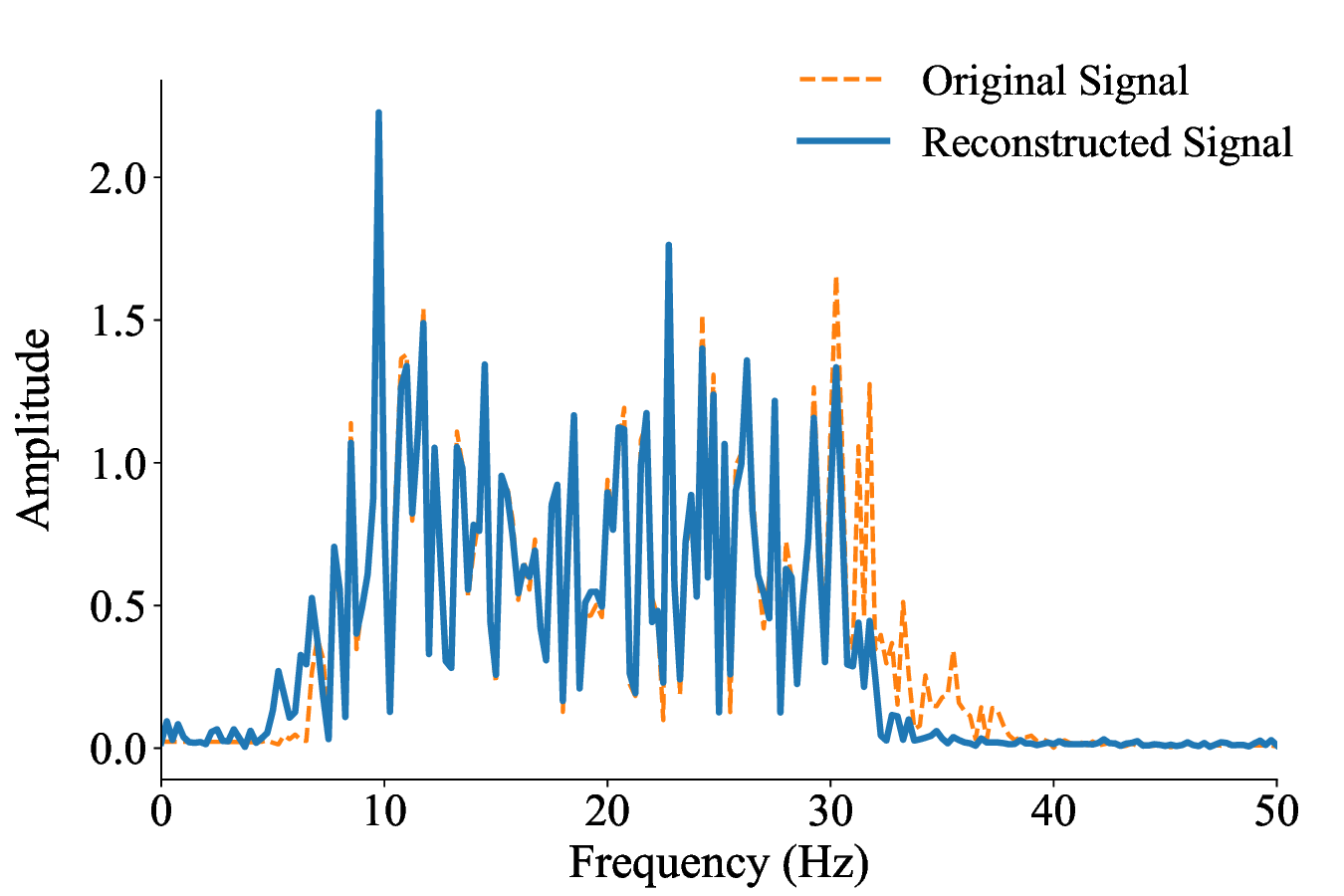}}
\caption{Frequency‐domain reconstruction under different compression ratios. (a) 5× compression; (b) 10× compression.}
\label{fig:spectrum_ratio}
\end{figure}

We revisit the compression ratio experiment to assess its effect on the learned latent space. Figure~\ref{fig:spectrum_ratio} shows the reconstructed signal spectra at 5× and 10× compression. While 5× compression preserves both low- and high-frequency EEG components, 10× compression severely attenuates higher-frequency bands, degrading signal fidelity. 

Results indicate that a moderate compression ratio (5×) and a feature selection rate in the range $[0.1,0.5]$ strike the best balance between diversity injection and preservation of target‐specific EEG features.  
\section{Conclusion}
\label{sec:conclusion}

This paper proposed FusionGen, a novel feature fusion-based EEG data generation framework for addressing data scarcity and inter-subject variability in BCIs. Existing augmentation methods either lack diversity or require large datasets, leading to suboptimal performance in few-shot scenarios. FusionGen employs disentangled representation learning to capture essential features and integrates them through a fusion module, enhancing data diversity while preserving physiological realism. Extensive experiments on various EEG datasets and paradigms demonstrate that FusionGen achieves superior performance in cross-subject and within-subject scenarios. FusionGen leverage disentangled representation learning for EEG data generation in BCIs, may serve as a powerful data engine for BCI large models. 
{
    \small
    \bibliography{main}
}



\end{document}